# AN IMAGE COMPRESSION AND ENCRYPTION SCHEME BASED ON DEEP LEARNING

**Fei Hu[1,2], Changjiu Pu[2], Haowei Gao[3], Mengzi Tang[1] and Li Li[1]**

[1] School of Computer and Information Science, Southwest University, Chongqing, China

[2] Network Centre, Chongqing University of Education, Chongqing, China

[3] The Webb Schools, 1175 West Baseline Road Claremont, CA 91711, USA

**ABSTRACT**

Stacked Auto-Encoder (SAE) is a kind of deep learning algorithm for unsupervised learning. Which has multilayers that project the vector representation of input data into a lower vector space. These projection vectors are dense representations of the input data. As a result, SAE can be used for image compression. Using chaotic logistic map, the compression ones can further be encrypted. In this study, an application of image compression and encryption is suggested using SAE and chaotic logistic map. Experiments show that this application is feasible and effective. It can be used for image transmission and image protection on internet simultaneously.

**Keywords:** Stacked Auto-Encode, deep learning, image protection, image feature, image compression, image encryption.

## 1. INTRODUCTION

With the development of multimedia technology and communication technology, multimedia entertainment has played an important role in people's daily lives. Pictures and videos take up the main part of multimedia entertainment. It brings austere challenge to store and transmit those data, and puts forward higher requirement on the limited-bandwidth internet, especially for large and high-quality digital images. The limited bandwidth of internet greatly restricts the development of image communication, and thus the image compression technology has been more and more aroused people's attention [1]. The purpose of image compression is to represent and transmit the original large image with minimal bytes, and to restore the image with not-so-bad quality. Image compression reduced the burden of image storage and transmission on the network, and achieved rapid real-time processing on line. The information of an image is fixed, but the different representations of the image lead to different changes in the amount of data stored in the image. So in the representation with larger amount of data, some data is useless or represent the information that is represented by other data, they are irrelevant or redundant. The main purpose of image compression is to compress the image by removing redundant or irrelevant information, and to store and transmit digital compressed data on a low bandwidth network.

Image compression techniques can be traced back to the digital television signals proposed in the year of 1948. There is almost 70 years of history. During this period there have been a variety of image compression coding methods. Especially in the late 1980s, due to the wavelet transform theory, the fractal theory, the artificial neural network theory and the visual simulation method, image compression technology was well developed. Image compression methods can be classified into two kinds: one may lose information during compression, and the other one can keep full information, that is, lossless coding methods and limited-distortion coding methods. Lossless coding methods will not suffer loss of information after compressing images, yet without a good compression ratio. The basic principle of this kind of methods is: an image consists of features, using the statistical features of the image, if a feature appears many times in the image, it will be encoded in shorter bits, and if a feature appears only once or limited times, it will be encoded in longer bits. And a complete image is always composed of a large number of repeated features. According to that, the image will be represented by many short-bits coding features and little long-bits coding features. On the basis of guaranteeing the image quality after compression, limited-distortion coding methods maximize the compression ratio. The original image and the compressed image looks very similar though some information has changed. The normal used limited-distortion coding methods are: the predictive coding method, the transform coding method and the statistical coding method. The limited-distortion coding method is more frequently uses than the lossless coding method because the former one has a larger compression ratio. Guaranteed under premise visual effects, which remove the information that the human eyes are not sensitive to.

The features of images can be learned automatically using deep learning models, rather than proposed manually. Suitable features can improve the performance of image recognition. Over the past years, features of images were always specified manually that depended on the designers' prior knowledge, and the number of features were very limited. Deep learning models can automatically learn unlimited number of features automatically. A good feature-extraction method is a prerequisite for optimization of image processing. Using deep learning models, unpredictable features of images can be learned, and these unpredictable features can also be used for image protection. In this study, we proposed a model to compress and encrypt images. Based on SAE, a multi-layer model is constructed. An image is put into the first layer and the output data from different level of layers reconstruct the original image in different level of comprehension. If the size of the output data from an arbitrary layer is less than the size of the original image, the representation in this layer is a compression representation. Because the model has more than one hidden layer whose neurons are less than the input layer's, the model can achieve multiple levels of features, and each level of features represent a compressed image. So, multiple compression ratio can be obtained using this model. The compressed image is further encrypted using chaotic logistic map. This model can be used in tasks that have certain requirements for image transmission speed and security.

## 2. RELATED WORKS
### 2.1. Stacked Auto-Encoder

Auto-Encoder (AE) is a single hidden layer model, and is an unsupervised learning neural network, see Fig. 1 (b). It is actually generated by two identical Restricted Boltzmann Machine models (RBMs) [2], see Fig. 1 (a). A RBM and a reversed RBM generate an AE model. Stacked Auto-Encoder (SAE) [3] is a multilayer AE, it is composed with several AEs. The previous AE's output is the later AE's input, i.e. several AEs' encoding sections are put together one by one and their decoding sections are put together in reverse order. This is a more complex AE model, having several hidden layers rather than one. Using greedy training methods, monolayer AE can be trained to learn weights directly, however, it is hard for SAE because more several hidden layers would consume too more computation time. In order to cut down the training time, the training process of SAE is divided into two steps: pre-training and fine-training. At first, each hidden layer is trained one by one, then the entire model is trained using the back propagation algorithm (BP).

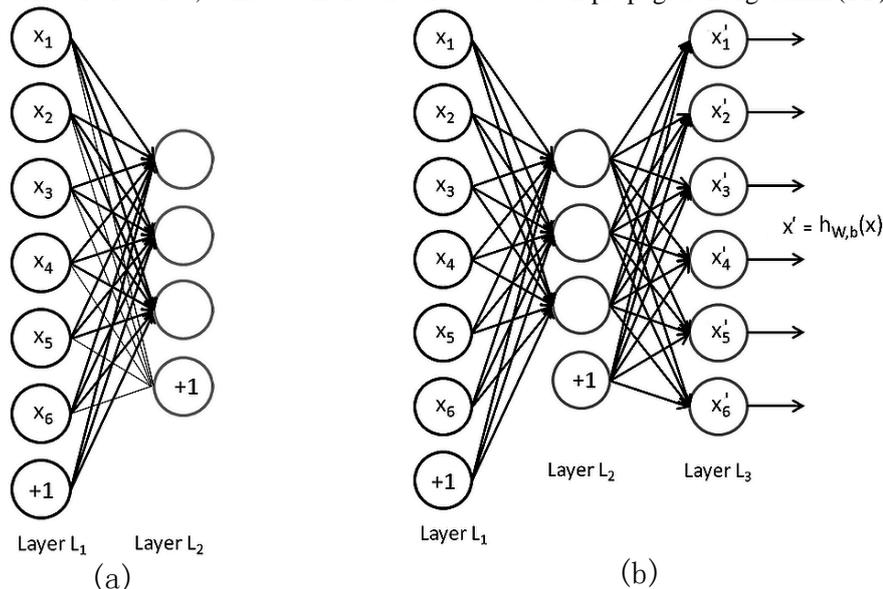

*Fig.1. RBM & Auto-encoder, (a) Restricted Boltzmann Machine (RBM), (b) Auto-Encoder.*

In Fig. 2, the left three layers （X, h1, h2） constitute the encoding part of the SAE model. In the pre-training phase, the input data X is encoded and yield h1, and then h1 is decoded and yield X', the error e=X'-X, e is used to adjust the weights between the layer X and the layer h1; then the output value of previous AE is set as the input data of layer h1, it is encoded and yield h2, h2 is decoded and yield h1', the weights between the layer h1 and the layer h2 are adjusted using e=h1'-h1; after multiple encoding and decoding operations, pre-optimal parameters (W, b) are got, and they make the model easy to train in the fine-training phase.

The encoding part （X, h1, h2） of the SAE model is flipped to get a decoding part: （h2, h1$^T$, X$^T$）. The two parts are combined to form a model that has the functions of encoding and decoding, see Fig. 2. In the fine-training phase, the weights are finely adjusted so that the optimal solution is closer little by little. Using the BP algorithm and the gradient descent algorithm, the fine-tuning process gradually approximates the optimal solution of the model. The detailed fine-training process is described as the following steps [4]:

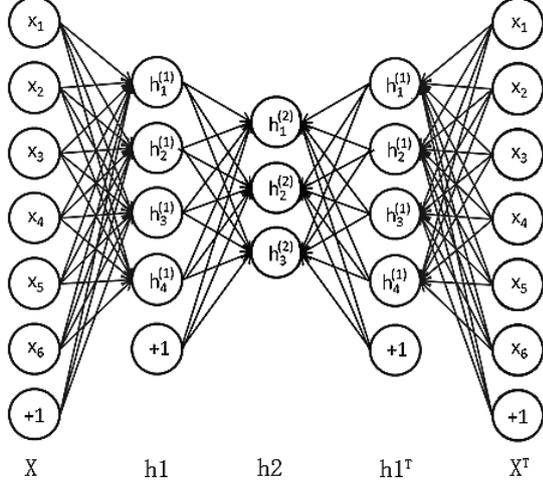

*Fig.2. Stacked Auto-encoder.*

1) Feedforward processing is performed by computing the activation for each middle layer;
2) The grade error for the output layer is
$$\delta^{(ni)} = -(y - a^{(ni)}) \cdot f'(z^{(ni)}) \quad (1)$$
3) The grade error for every middle layer is
$$\delta^{(l)} = ((W^{(l)})^T \delta^{(l+1)}) \cdot f'(z^{(l)}) \quad (2)$$
4) The partial derivatives:
$$\nabla_{W^{(l)}} J(W, b; x, y) = \delta^{(l+1)} (a^{(l)})^T \quad (3)$$
$$\nabla_{b^{(l)}} J(W, b; x, y) = \delta^{(l+1)} \quad (4)$$
5) the overall cost function is set as the following:
$$J(W, b) = \left[ \frac{1}{m} \sum_{i=1}^{m} J(W, b; x^{(i)}, y^{(i)}) \right] \quad (5)$$

SAE has strong representation expression ability and advantages of deep neural networks. AE can learn the characteristics of input data, then SAE can learn multi-level characteristics. In the first hidden layer, SAE can learn first-order features of the input data; in the second hidden layer, SAE can learn second-order features of the input data. E.g. the input data is a set of images, the first hidden layer may learn a collection of edges, and the second hidden layers may learn how to combine a number of edges together to form an outline, a higher hidden layer may learn much more vivid, special and meaningful features. Features of each level can help us better operate image processing, such as image classification, information retrieval of images, and so on. These features also can be used to compress images. For example, an image with 100 pixels is put into the input layer, the input layer has 100 neurons, each pixel is put into a corresponding neuron, then a hidden layer with only 10 neurons yields a 10-dimensional vector, which owns features of the input data and can be considered as a reconstruction of the input image, so this image is compressed and the compression ratio is 10.

**2.2. Image encryption schema using chaotic logistic map**

Chaos-based cryptographic algorithms have suggested efficient ways to develop secure image encryption. These algorithms are sensitive to their initial conditions. Any tiny change can cause greatly different responses that guarantees the efficiency of encryption schemas. The logistic map is one of them. It is an iterated logistic map that has proved great importance in many fields of information processing. Such fields include but are not limited in the following: population biology, chemistry, encryption, communication and ecology. It also works in modelling the dynamics of a single species. The stability and bifurcation of the logistic map has been studied a lot, such as Cohen-Grossberg neural networks with delays [5] and the Neimark-Sacker bifurcation with delay [6].

The logistic map is of a non-linear recursive relation. It can suggest deterministic chaos. Its mathematical equation is written as:
$$x_{n+1} = r x_n (1 - x_n) \quad (6)$$
where $x_0$ is an initial condition which is a float number between 0 and 1 (exclude 0 or 1), r is a positive constant which is also a float number between 3.5699 and 4 (include 4, and 3.5699 is an approximation). After N iterations, a sequence will be got. The sequence is like the form of $\{x_1, x_2, x_3, \ldots, x_N\}$. It is a stochastic sequence which can further be used for encryption tasks. In this study, the initialized number $x_0$ is generated from the SAE model, and then is used for image encryption.

**3. IMAGE COMPRESSION ENCRYPTION MODEL**

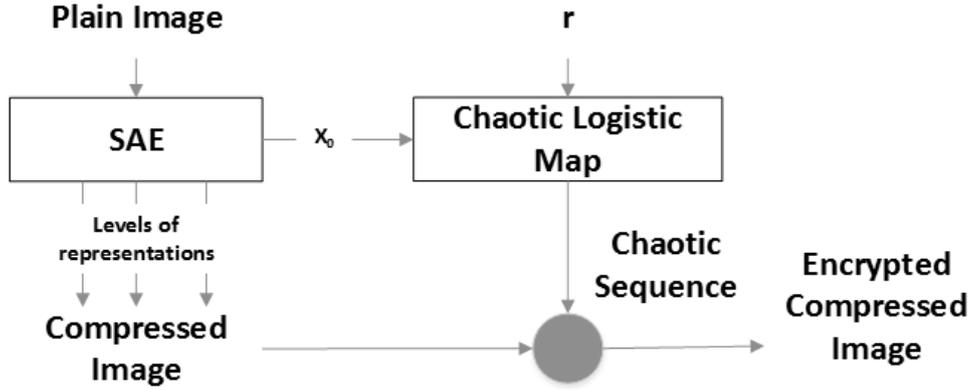

*Fig.3. The compression encryption model.*

The image is compressed using the SAE, and then the compressed image is encrypted using the sequence generated by the chaotic logistic map.

The diagram in Fig. 3 shows the algorithm of this model. See the following for detail:

1) Initialization:

A five-layer SAE model is constructed (see Fig. 2). In the model, the second layer has a less number of neurons than the input layer in order to realize the primary image compression, and the third layer has a less number of neurons than the second layer in order to realize the second-stage image compression. The rest fourth and fifth layers are separately the mirrors for the second and first layers. In the study of image processing using Convolutional Neural Networks (CNN), it was supposed and has been proved by a large number of experiments [7] that an image could be divided into a number of regions and the characteristics learned by CNN from different regions are similar or even the same. We use this supposition in our study. The image is divided into pieces which have the same size. Every piece is a sample. A training set consists of all pieces from one single image. For the convenience of handling images in SAE, values in this data set will be normalized as float numbers in the range of 0 to 1 before they are put into the model. Our model is trained using this normalized data set. Levels of learned features will be anti-normalized to get the final output, which are values of pixels for a compressed image. According to the dense representation, the image is compressed. The constant r is initialized;

2) Learning compression representation using the SAE model:

The activation function f (·) is a nonlinear function, and the sigmoid function is used in the experiment. By training the SAE model, we get a compression representation from an arbitrary hidden layer. This representation then forms a compressed image. Because the sigmoid function output is a float number between 0 and 1, which meets the requirements for initializing $x_0$, a certain one from the output values is chosen as $x_0$. In the experiment, the first one is chosen;

3) Generating a chaotic sequence:

Using chaotic logistic map with $x_0$ and r, a sequence S is generated, S= {$x_1$, $x_2$, $x_3$, …, $x_N$}, where N is the size of the compressed image, e g, a compressed image has 100*100 pixels, N=100*100=10000;

4) Image encryption and image decryption:

The encryption and decryption functions are described following, where E is the encrypted image, C is the plain image (the compressed image), S is the sequence generated in step 3), bitxor(-) is a bit XOR function;

$$E = bitxor(C,S) \quad (7)$$
$$C = bitxor(E,S) \quad (8)$$

5) Image reconstruction:

The compressed image is recovered through the SAE model. See Fig. 2, if the compressed image came from the layer of h2, a new model is reconstructed with only the layers of {X, h1, h2} which shares the parameters learned in step 2). The compressed image is normalized and is put into the layer of h2, then the output from the layer of X represent the recovered image.

The practicability of the algorithm will be verified in the next section.

## 4. EXPERIMENTS

This new model was evaluated on several images taken from the standard set of images. They are house, airplane, lake and pepper. They have the same size as 512 by 512. Images split into pieces with the same size of 8 by 8. The number of neurons in the input layer was 64, and the number of neurons in the hidden layers was adjusted to achieve different compression ratios (CRs), that is, 4:1 and 16:1 for 16 and 4 neurons. In the back part of this section, the compressed images were encrypted. Correlation Analysis was performed to evaluate the effect of the encryption schema.

Compression effects are shown in Fig. 4. In order to quantitatively verify the effects, Mean Square Error (MSE) and Peak Signal-to-Noise

Ratio (PSNR) are introduced. MSE [8] is the average of the square of the difference between the expected response and the actual output. It is also called squared error loss. PSNR [9] is the ratio of maximum power of the signal and the power of noise. It is commonly used to measure the quality of reconstruction in image compression. Their mathematical definitions are following equations:

$$\text{MSE} = \frac{1}{m \times n} \sum_{i=0}^{m-1} \sum_{j=0}^{n-1} [I_O(i,j) - I_R(i,j)]^2 \quad (9)$$

$$PSNR = 10 \log_{10}(\frac{MAX_I^2}{MSE}) \quad (10)$$

where $MAX_I$ is the maximum possible pixel value of the image, that is 255 in this experiment, m*n is the image size, $I_O$ is the original image and $I_R$ is the reconstructed image.

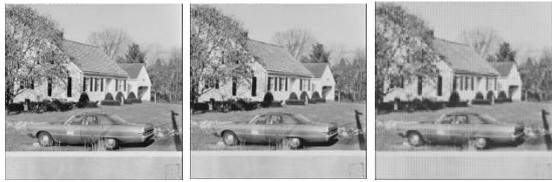

(a) Original image of house, image reconstructed at a CR of 4:1, image reconstructed at a CR of 16:1

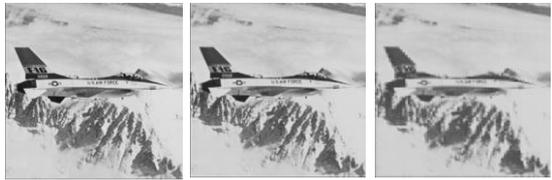

(b) Original image of airplane, image reconstructed at a CR of 4:1, image reconstructed at a CR of 16:1

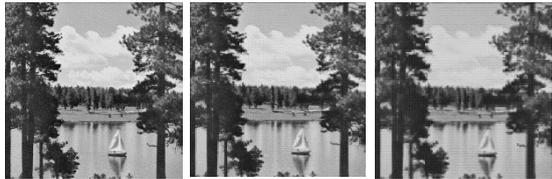

(c) Original image of lake, image reconstructed at a CR of 4:1, image reconstructed at a CR of 16:1

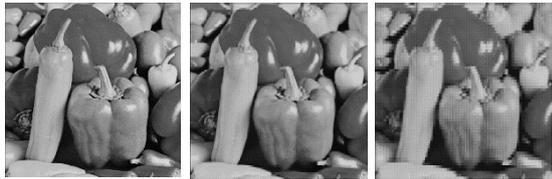

(d) Original image of pepper, image reconstructed at a CR of 4:1, image reconstructed at a CR of 16:1

*Fig.4. Compression effects.*

MSE and PSNR were computed for three primary colour channels (Red, Green and Blue, also called RGB), respectively. And the results at the CRs of 4:1 and 16:1 are listed in Table 1 and Table 2, respectively.

The encrypted images of compression ones at CRs of 4:1 and 16:1 are shown in Fig. 5 and Fig. 6, respectively. Correlation Analysis was performed to quantitatively evaluate the effect of the encryption schema. The correlation coefficient is used to evaluate the correlation of a pair of adjacent pixels, and it is defined below [10]:

$$r_{xy} = (E(xy) - E(x)E(y))/(\sqrt{D(x)}\sqrt{D(y)}) \quad (11)$$

where $r_{xy}$ is the correlation coefficient of the variables x and y, E (·) is the mean function, D (·) is the variance function, x and y are adjacent pixels.

*Table 1.* MSE and PSNR of the reconstructed images at a CR of 4:1.

| Images | | R | G | B |
|---|---|---|---|---|
| House | MSE | 37.7696 | 51.1280 | 37.0831 |
| | PSNR | 32.3594 | 31.0442 | 32.4390 |
| Airplane | MSE | 21.4800 | 32.9829 | 15.8055 |
| | PSNR | 34.8105 | 32.9479 | 36.1427 |
| Lake | MSE | 32.2390 | 101.1015 | 56.4297 |
| | PSNR | 33.0470 | 28.0832 | 30.6157 |
| Pepper | MSE | 48.8470 | 60.6932 | 39.5987 |
| | PSNR | 31.2424 | 30.2994 | 32.1540 |

*Table 2.* MSE and PSNR of the reconstructed images at a CR of 16:1.

| Images | | R | G | B |
|---|---|---|---|---|
| House | MSE | 110.9180 | 136.8415 | 118.2971 |
| | PSNR | 27.6808 | 26.7686 | 27.4011 |
| Airplane | MSE | 119.2817 | 114.3715 | 71.0135 |
| | PSNR | 27.3651 | 27.5476 | 29.6174 |
| Lake | MSE | 62.2023 | 174.2972 | 144.9320 |
| | PSNR | 30.1927 | 25.7179 | 26.5192 |
| Pepper | MSE | 71.5491 | 149.3480 | 74.9371 |
| | PSNR | 29.5848 | 26.3888 | 29.3838 |

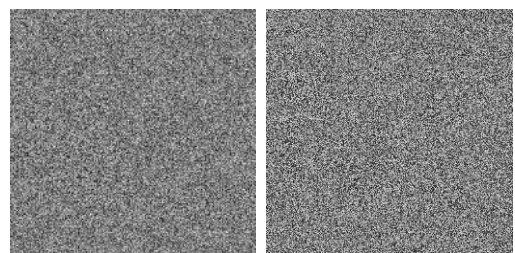

(a) (b)

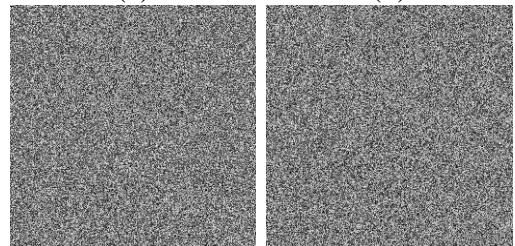

(c) (d)

*Fig.5. encrypted images of compression ones at a CR of 4:1, (a) encrypted house, (b) encrypted airplane, (c) encrypted lake, (d) encrypted pepper.*

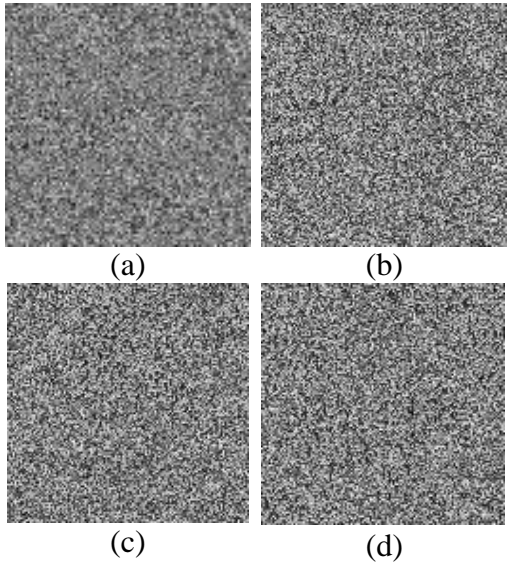

(a)      (b)

(c)      (d)

*Fig.6. encrypted images of compression ones at a CR of 16:1, (a) encrypted house, (b) encrypted airplane, (c) encrypted lake, (d) encrypted pepper.*

To simplify the processing, colour images were converted to grayscale before use. Experiments were repeated ten times, correlation coefficients of each image were averaged. Results for reconstructed images and encrypted compression images at the CRs of 4:1 and 16:1 are listed in Table 3.

*Table 3.* Correlation Analysis for reconstructed images and encrypted compression images at the CRs of 4:1.

| Images | CR of 4:1 | | CR of 16:1 | |
| --- | --- | --- | --- | --- |
| | Compressed | Encrypted | Compressed | Encrypted |
| House | 0.9532 | -0.0391 | 0.9369 | 0.0079 |
| Airplane | 0.9632 | 0.0038 | 0.9745 | 0.0361 |
| Lake | 0.9820 | -0.0087 | 0.9855 | 0.0267 |
| Pepper | 0.9796 | 0.0488 | 0.9902 | 0.0261 |

The experimental results show that this new model is effective and it can be used for image transmission and image protection on internet simultaneously.

## 5. CONCLUSIONS

A scheme of deep learning in image compression and encryption was proposed. Based on SAE neural networks, images are compressed. And then the compressed ones are encrypted using chaotic logistic map. This scheme can be used for image transmission and image protection on internet simultaneously.

**Acknowledgements**

This work was supported by Scientific and Technological Research Program of Chongqing Municipal Education Commission (No. KJ1501405, No. KJ1501409); Scientific and Technological Research Program of Chongqing University of Education (No. KY201522B, No. KY201520B); Fundamental Research Funds for the Central Universities (No. XDJK2016E068); Natural Science Foundation of China (No. 61170192) and National High-tech R&D Program (No. 2013AA013801).